\documentclass{article}
\usepackage{spconf,amsmath,graphicx}

\usepackage{times}
\usepackage{epsfig}
\usepackage{graphicx}
\usepackage{amsmath}
\usepackage{amssymb}
\usepackage{booktabs}
\usepackage{lipsum}
\usepackage{color}

\usepackage{enumitem}
\setlist{nosep, leftmargin=14pt}

\usepackage{mwe} 

\title{Cluster-Guided Semi-Supervised Domain Adaptation for Imbalanced Medical Image Classification}
\name{
Shota Harada$^{1}$,
Ryoma Bise$^{1}$,
Kengo Araki$^{1}$,
Akihiko Yoshizawa$^{2}$,
Kazuhiro Terada$^{2}$}
\secondlinename{
Mariyo Kurata$^{2}$,
Naoki Nakajima$^{2}$,
Hiroyuki Abe$^{3}$,
Tetsuo Ushiku$^{3}$,
Seiichi Uchida$^{1}$
}
\address{
$^{1}$ Kyushu University, Fukuoka, Japan, 
$^{2}$ Kyoto University, Kyoto, Japan, \\
$^{3}$ The University of Tokyo, Tokyo, Japan
}
\begin{document}
%
\maketitle
\begin{abstract}
Semi-supervised domain adaptation is a technique to build a classifier for a target domain by modifying a classifier in another (source) domain using many unlabeled samples and a small number of labeled samples from the target domain. In this paper, we develop a semi-supervised domain adaptation method, which has robustness to class-imbalanced situations, which are common in medical image classification tasks. For robustness, we propose a weakly-supervised clustering pipeline to obtain high-purity clusters and utilize the clusters in representation learning for domain adaptation. The proposed method showed state-of-the-art performance in the experiment using severely class-imbalanced pathological image patches.
\end{abstract}
%
\section{Introduction}

\begin{figure*}[t]
\center
\includegraphics[width=\textwidth]{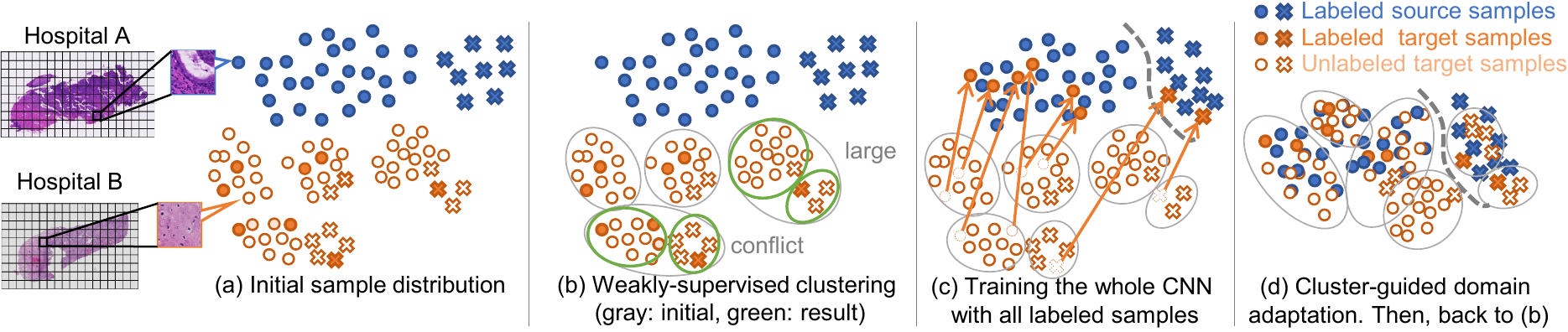}
\vspace{-.8cm}
\caption{The proposed semi-supervised domain adaptation method. This figure assumes a two-class task ($\circ$ and $\times$) for simplicity, although the method can deal with a $C$-class task ($C \geq 2$). (a) Initial sample distribution in semi-supervised domain adaptation scenario. (b)~Weakly-supervised clustering of target domain samples. Two refinement steps are essential to increase cluster purity. (c)-(d)~The main domain adaptation steps using the clustering result. The steps (b)-(d) are repeated until convergence.}
\label{fig:overall}
\vspace{-.2cm}
\end{figure*}
Domain adaptation is an important technique to build a classifier for a {\em target} domain while utilizing the labeled samples of a {\em source} domain~\cite{Long2015icml,Yaroslav2015icml,tzeng2017cvpr,wang2021pr,Shuhan2020}. In a typical {\em unsupervised} domain adaptation scenario, all samples in the target domain are unlabeled. Then, by modifying the classifier using the unlabeled samples, the classifier becomes working for the target domain samples. The typical ``domain'' in the medical image analysis is ``hospital.'' For example, as shown in the left of Fig.~\ref{fig:overall}, in pathological image analysis, whole slide images (WSIs) taken at different hospitals often show different appearance due to difference in microscopes, staining, and scanning device.
\par
{\em Semi-supervised} domain adaptation is another scenario~\cite{saito2019semi,Pin2020ijcai,He2020cvprws,li2021cross}. As shown in Fig.~\ref{fig:overall}(a), we have a small number of labeled samples even in the target domain. Using them as bridges between the two domains will have more reliable effects on domain adaptation than the unsupervised scenario. In medical applications, we can expect to have a small number of labeled samples from the target hospital, and therefore, the semi-supervised domain adaptation scenario is more feasible for medical image classification tasks (by paying small efforts).
\par
This paper proposes a cluster-guided semi-supervised domain adaptation method, as shown in Fig.~\ref{fig:overall}. It is natural to assume that the unlabeled samples from the target domain are expected to be distributed around the labeled target samples of the same class, as shown in Fig.~\ref{fig:overall}(a). Based on this assumption, we introduce a clustering procedure for all the labeled and unlabeled target samples. If we get clusters with high purity as (b), we can perform cluster-wise domain adaptation where the labeled samples help the unlabeled samples in the same cluster, as shown in (c) and (d).
\par
For the proposed method, it is crucial to have high-purity clusters. For medical images, however, naive clustering techniques are not appropriate. This is because medical images are often heavily class-imbalanced and naive clustering techniques result in low-purity clusters like the gray clusters of Fig.~\ref{fig:overall}(b). What is worse, this class-imbalanced problem has not been well considered in the past domain adaptation techniques, such as~\cite{Shuhan2020,saito2019semi,li2021cross}. As shown in the later experiments, they actually show poor performance in class-imbalanced classification tasks.
\par
We, therefore, propose a novel clustering technique, called {\em weakly-supervised clustering} to have clusters with high purity, like the green clusters of Fig.~\ref{fig:overall}~(b). In weakly-supervised clustering, we introduce two refinement steps for the initial clustering results by $k$-means. The first refinement step is cluster refinement by soft-constrained clustering. The second is cluster refinement by proportion-based splitting. We also propose an objective function for guiding unlabeled target samples belonging to the same cluster by labeled target samples.
\par
We evaluated the proposed method using two types of datasets: One type is digits datasets and the other is the WSI datasets taken at two hospitals for a cervical cancer stage classification. The experimental results show that the proposed method achieved the state-of-the-art (SOTA) performance and showed the expected robustness to a severely class-imbalance condition.
\par
The main contributions of this paper are summarized as follows.
\begin{itemize}
\item We propose a semi-supervised domain adaptation method that relies on a novel clustering pipeline, called weakly-supervised clustering, for higher purity even under a heavy class-imbalance condition. 
\item Experimental results under a realistic setup show the expected effects of the above proposals and achieve better performance than the existing methods.
\end{itemize}

\section{Cluster-guided semi-supervised domain adaptation}
In the proposed method,  we first perform a novel weakly-supervised clustering technique to have high-purity clusters even under a heavy class-imbalance condition. Then, we perform a novel cluster-guided domain adaptation technique, where the labeled target samples guide the unlabeled samples belonging to the same cluster, as shown in Fig.~\ref{fig:overall}(d). This adaptation strategy is reasonable with  high-purity clusters, which are provided by our weakly-supervised clustering. 
\par
\subsection{Problem setting}
Consider that we have a set of $m^{s}$ labeled source samples, $\mathcal{D}^{s}=\{(\boldsymbol{x}^{s}_{i}, y^{s}_{i})\}_{i=1}^{m^{s}}$, where $\boldsymbol{x}^{s}_{i}$ is the $i$-th image sample in the source domain and $y^{s}_{i}\in \{1,\ldots,C\}$ is its class label. In the target domain, we have a set of $m^{t}$ labeled samples, $\mathcal{D}^{t}=\{(\boldsymbol{x}^{t}_{i}, y^{t}_{i})\}_{i=1}^{m^{t}}$, and a set of $m^{u}$ unlabeled target samples, $\mathcal{D}^{u}=\{\boldsymbol{x}^{u}_{i}\}_{i=1}^{m^{u}}$. Then we consider the problem of improving the classification performance using not only $\mathcal{D}^{s}$ but also $\{\mathcal{D}^{t}, \mathcal{D}^{u}\}$, {\em after adapting $\{\mathcal{D}^{t}, \mathcal{D}^{u}\}$ to $\mathcal{D}^{s}$}. Since we have the labeled samples $\mathcal{D}^{t}$ in the target domain, this problem is called semi-supervised domain adaptation.
\par
\subsection{Weakly-supervised clustering}
\label{sec:clustering}
Our weakly-supervised clustering comprises three steps. As the first step, $k$-means clustering is performed for {\em target} samples $\{\mathcal{D}^{t}, \mathcal{D}^{u}\}$, shown as the gray clusters of Fig.~\ref{fig:overall}(b). Here, the number of clusters $K$ is set to be larger than the actual number of classes in the dataset. Then, carefully-designed two cluster refinement steps are performed to improve the purity of clusters, even for a minor class, as shown by the green clusters in Fig.~\ref{fig:overall}(b). The two refinement steps are detailed in the following.
\par
\noindent{\bf Cluster refinement by soft-constrained clustering}:\ 
The first cluster refinement step aims to divide every conflicting cluster (i.e., a cluster with samples from different class labels) into several non-conflicting clusters. For this aim, we use ``soft''-constrained clustering~\cite{harada2021mia}. In general, constrained clustering introduces two types of constraints, called must-link and cannot-link. A must-link is given to a pair of samples that should be grouped into the same cluster, whereas a cannot-link is given to samples that should not be grouped.
\par
In our task, if we find samples with different labels in the same initial cluster (by $k$-means), cannot-links are attached to all pairs of those samples. Similarly, if we find samples with the same label in the initial cluster, must-links are attached to them. After the attachment of the links, we perform clustering again while satisfying the constraints of the links. Note again that we use ``soft''-constrained clustering. Since ordinary constrained clustering, such as hard-constrained clustering, may cause low-purity clusters due to a must-link for distant samples, we use soft-constrained clustering, which allows the violation of such must-links. After applying this step, labeled target samples in a cluster always belong to a single class.
\par
\noindent{\bf Cluster refinement by proportion-based splitting}:\
The second cluster refinement step aims to split a cluster into smaller clusters according to class proportions $(p_1,\ldots,p_C)$, (i.e., prior class probability), which is inferred by the class ratio among the labeled target samples. This aim is similar to the previous refinement step but uses a different criterion. Roughly speaking, after the first refinement, if we find a large cluster that contains one or more labeled samples, the cluster will be a non-pure cluster and should be split into smaller clusters.\par
More specifically, we split the larger clusters into smaller clusters, by using class proportions. Let $\bar{c}_i$ denote the class of the labeled samples in the $i$-th cluster, and $u_i$ is the number of unlabeled samples in the $i$-th cluster. Then, if $m^{u} p_{\bar{c}_i} \leq {u_i}$, we consider that the cluster is too large for the class $\bar{c}_i$ and thus divide it into two smaller clusters by $k$-means ($k=2$). Consequently, we can expect high-purity clusters even for minor classes. 
\par
\subsection{Cluster-guided domain adaptation}
Using the clustering result given by the above careful steps, we now perform cluster-guided domain adaptation, as shown in Figs.~\ref{fig:overall}(c) and (d). The CNN model $\boldsymbol{f}$ is trained for two objectives. One is the classification of all labeled samples, $\mathcal{D}^{s} \cup \mathcal{D}^{t}$, by the cross-entropy loss to bring the source sample and the labeled target sample closer together, as shown in Fig.~\ref{fig:overall}(c). The other is to guide the unlabeled sample $\boldsymbol{x}^{u}_{j}$ closer to the labeled sample $\boldsymbol{x}^{t}_{i}$ that belongs to the same cluster after re-training (i.e., $\boldsymbol{x}^{t}_{i}$ is closer to $\boldsymbol{x}^{u}_{j}$ that belongs to the same cluster than $\boldsymbol{x}^{u}_{l}$ that belongs to a different cluster). More specifically, we train the model by the following objective:
\begin{flalign}
\label{eq:cluster-loss}
 &\mathcal{L}_{\textrm{clu}}(\boldsymbol{x}^{t}_{i}, \boldsymbol{x}^{u}_{j}, \boldsymbol{x}^{u}_{l})\!= \\
 &\!\max\left\{ ||\boldsymbol{f}(\boldsymbol{x}^{t}_{i})-\boldsymbol{f}(\boldsymbol{x}^{u}_{j})||^2_2\!-\!||\boldsymbol{f}(\boldsymbol{x}^{t}_{i})-\boldsymbol{f}(\boldsymbol{x}^{u}_{l})||^2_2\!+\! \varepsilon,\ 0 \right\}, \nonumber
\end{flalign}
where $\boldsymbol{f}(\boldsymbol{x})$ denotes the feature vector for the sample $\boldsymbol{x}$ and $\varepsilon \in \Re^+$ is the margin. As shown in Fig.~\ref{fig:overall}(d), the unlabeled samples are gradually mapped to the corresponding class of the source domain by training $\boldsymbol{f}$ with this loss, and the guidance of the labeled sample $\boldsymbol{x}^{t}_{i}$. Note that we do not give any pseudo-label to the unlabeled samples in this framework -- the unlabeled samples are used as unlabeled for helping the representation learning with the labeled samples in Eq.~(\ref{eq:cluster-loss}).
\par
\section{Experiments}
\subsection{Experimental setup}
We conducted experiments using two types of datasets: 1) digits datasets, which is one of the famous domain adaptation setting \cite{Yaroslav2015icml}, and 2) cervical cancer dataset, which was collected from several hospitals, as real clinical data. 
\par
\noindent{\bf Digits datasets}:\ 
We used three digits datasets: MNIST~\cite{Lecun1998}, USPS~\cite{Hull1994}, and SVHN~\cite{Netzer2011}, which are used in the paper on class-imbalanced domain adaptation~\cite{Shuhan2020}. To simulate class imbalance problems, which often occur in medical image analysis, we undersampled the original data according to the Pareto distribution~\cite{William2001}, which is one of a heavy-tailed distribution. We evaluated our method using pairs of these datasets as the source and target domains. All images were labeled in the source domain, and 2$\%$ images were randomly sampled as labeled data in the target domain.
\par
\noindent{\bf Cervical cancer dataset}:\ 
We collected WSI images from two hospitals and used them as the source and target domains. The source domain dataset contains $158$ WSIs ($\mathcal{D}^s$), and the target domain dataset contains $106$ WSIs ($\mathcal{D}^t + \mathcal{D}^u$). Each WSI was cut into patches of $256\times256$ pixels as shown in the left of Fig.~\ref{fig:overall}.
Consequently, $163,877$ patch images were collected from the source domain and $71,598$ from the target domain. In the experiments, 5-fold cross-validation was conducted with patient-disjoint random sample splitting, where the same patient was not present in the training and test sets.
\par
Each sample was labeled as one of three cervical cancer stages: Non-Neoplasm (Non-Neop.), Low-Grade Squamous Intraepithelial Lesion (LSIL), and High-Grade Squamous Intraepithelial Lesion (HSIL). This dataset shows a heavy class imbalance among the three classes. For example, the samples of the source domains are $146,524$ Non-Neop., $5,646$ LSIL, and $11,707$ HSIL. 
\par
\begin{figure*}[t]
\center
\includegraphics[width=\textwidth]{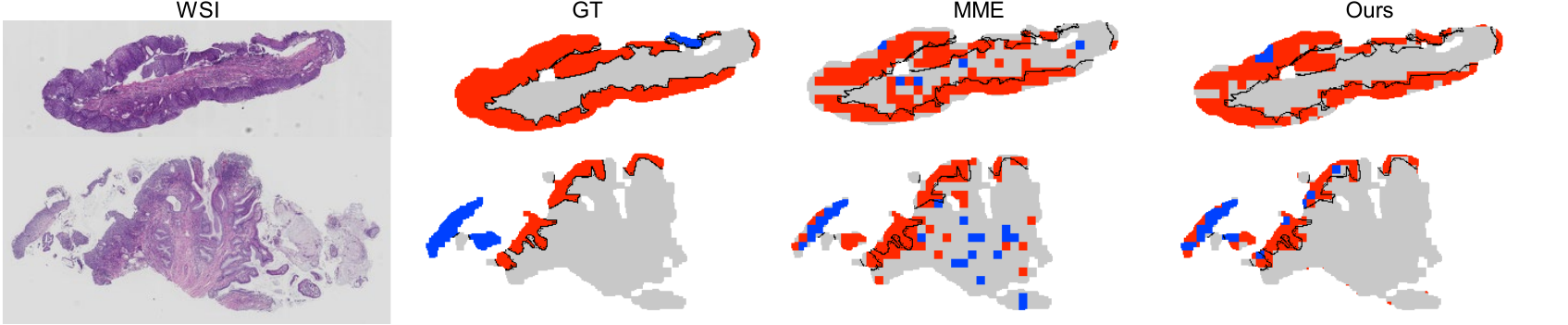}
\vspace{-.8cm}
\caption{Examples of segmentation results from the target domain. The columns from the right show the segmentation results of MME~(as the second best in Table~\ref{tab:performance}) and the proposed method. Gray, blue, red, and black indicate Non-Neop., LSIL, HSIL, and Nolabel area, respectively.} 
\label{fig:predict-examples}
\vspace{-.2cm}
\end{figure*}
\noindent{\bf Comparative methods}:\
We compared the proposed method with seven methods, including current SOTA methods. Two methods are simple supervised learning methods: One is a model trained with the labeled source samples~(\textbf{S}), and the other is a model trained with the labeled source and labeled target samples~(\textbf{S+T}). \textbf{DANN}~\cite{Yaroslav2015icml} and \textbf{COAL}~\cite{Shuhan2020} were unsupervised domain adaptation methods,
and \textbf{ENT}~\cite{yve2004nips}, \textbf{MME}~\cite{saito2019semi}, and \textbf{CDAC}~\cite{li2021cross} were semi-supervised domain adaptation methods. For a fair comparison, we modified these unsupervised domain adaptation methods to adapt to semi-supervised domain adaptation by using $\mathcal{D}^{t}$ for supervised learning in their methods. In addition, an ablation study was conducted using the digits datasets to evaluate the effectiveness of each step of our method.
\par
\noindent{\bf Evaluation metrics}:\ 
We used the mean of the per-class Dice-coefficient (mDice), which has been widely used for segmentation performance. To show the effectiveness of our method for imbalanced data, we also showed the Dice coefficient for the minor class, the LSIL class (MiDice), in the experiment using the cervical cancer dataset.
\par
\noindent{\bf Implementation details}:\
We first initialized ResNet-50~\cite{kaiming2016CVPR}, which was pretrained using ImageNet~\cite{olga2015ijcv}, as a model $\boldsymbol{f}$ by training it using the labeled source domain data $\mathcal{D}^{s}$. We then retrained the model by repeating the steps of Fig.~\ref{fig:overall}(b)-(d). We terminated the training of the model by early stopping referred to mDice of the validation set. The number of clusters $K$ for the initial clustering ($k$-means) was set to $30$, which is larger than the number of classes in both datasets. In addition, the number of clusters for soft-constrained clustering was set to $K$ multiplied by the ratio of target samples belonging to that conflicting cluster to all target samples.
\par
\begin{table}[t]
\caption{Mean of the per-class Dice-coefficient (mDice) on the digits datasets.}
\label{tab:digit-performance}
\centering
\resizebox{\columnwidth}{!}{
\begin{tabular}[t]{l c c c c c}
\hline
Method & 
\begin{tabular}[c]{c}
MNIST\\$\rightarrow$USPS
\end{tabular}
&
\begin{tabular}[c]{c}
USPS\\$\rightarrow$MNIST
\end{tabular}
& 
\begin{tabular}[c]{c}
SVHN\\$\rightarrow$MNIST
\end{tabular}
& 
\begin{tabular}[c]{c}
SVHN\\$\rightarrow$USPS
\end{tabular}
& 
Avg\\
\hline
\hline
S & $0.5213$ & $0.4821$ & $0.4457$ & $0.6277$ & $0.5192$  \\
S+T & $0.6528$ & $0.5420$ & $0.6783$ & $\mathbf{0.6607}$ & $0.6334$ \\
DANN~\cite{Yaroslav2015icml} & $0.7062$ & $0.4987$ & $0.6685$ & $\mathbf{0.6607}$ & $0.6335$ \\
ENT~\cite{yve2004nips} & $0.7326$ & $0.4549$ & $0.6654$ & $0.5239$ & $0.5942$ \\
MME~\cite{saito2019semi} & $0.7314$ & $0.4734$ & $0.6496$ & $0.4982$ & $0.5881$ \\
COAL~\cite{Shuhan2020} & $\mathbf{0.7521}$ & $0.7555$ & $0.6451$ & $0.4725$ & $0.6563$ \\
CDAC~\cite{li2021cross} & $0.7765$ & $0.5932$ & $0.4315$ & $0.5575$ & $0.5897$ \\
Ours & $0.6862$ & $\mathbf{0.7996}$ & $\mathbf{0.6974}$ & $0.6564$ & $\mathbf{0.7094}$ \\
\hline
KM & $0.7589$ & $0.6973$ & $0.6002$ & $0.6376$ & $0.6735$ \\
Soft Const. & $0.7274$ & $0.5852$ & $0.6585$ & $0.6376$ & $0.6522$ \\
\hline
\end{tabular}
}
\end{table}

\begin{table}[t]
\caption{Results on the cervical cancer stage classification.}
\label{tab:performance}
\centering
\resizebox{\columnwidth}{!}{
\begin{tabular}[t]{l c c c c c c c c}
\hline
Method & S & S+T & DANN & ENT & MME & COAL & CDAC & Ours\\
\hline
\hline
mDice & $0.4740$ & $0.5617$ & $0.5128$ & $0.4969$ & $0.5051$ & $0.4673$ & $0.4710$ & $\mathbf{0.5897}$\\
\hline
MiDice & $0.1073$ & $0.1821$ & $0.1331$ & $0.1154$ & $0.1397$ & $0.1102$ & $0.0980$ & $\mathbf{0.1937}$\\
\hline
\end{tabular}
}
\end{table}
\subsection{Experimental results}
\noindent{\bf Digits datasets}:\
Table~\ref{tab:digit-performance} shows the mDice under several combinations of the domains. The proposed method outperformed the comparative methods in Avg, and only the proposed method improved the performance compared with S+T in all combinations. In contrast, most comparative methods could not improve the performance compared with S+T. The lack of improvement in the performance for digit classification tasks, a relatively simple task, shows that comparative methods are not robust to class imbalance. This result confirms the proposed method is robust to class imbalance.
\par
This result shows that the proposed method is not effective when the USPS is set as the target domain. Since the proposed method uses labeled target samples as an anchor, it is not expected to make a significant effect when the number of samples is extremely small, and the number of labeled target samples is limited; such as the USPS. Specifically, the number of samples in the MNIST, USPS, and SVHN were $8{,}396$, $836$, and $7{,}216$, respectively.
\par
\noindent{\bf Ablation study}:\
To evaluate the effectiveness of each step in the weakly-supervised clustering, we evaluated ablated methods: \textbf{KM} is a method that used $k$-means for cluster-guided domain adaptation (i.e., without soft-constrained clustering and proportion-based splitting); \textbf{Soft Const.} was using the clusters refined by soft-constrained clustering (i.e., without proportion-based splitting). As shown in the bottom of Table~\ref{tab:digit-performance}, the results demonstrate that each refinement step effectively improves classification performance.
\par
\noindent{\bf Cervical cancer dataset}:\
Table~\ref{tab:performance} shows the performance of each method using the Cervical cancer datasets. All the comparative methods had decreased performance compared with S+T as a baseline. In contrast, the proposed method achieved improving performance and outperformed the comparative methods in MiDice. These results confirm that the proposed method is robust to the class-imbalanced scenario, which is a typical situation of medical image classification.
\par
Fig.~\ref{fig:predict-examples} shows the examples of segmentation results of the target domain WSIs obtained from MME, which is the second-best in Table~\ref{tab:performance}, and the proposed method. Fig.~\ref{fig:predict-examples} confirmed that MME mispredicts the Non-Neop. to the minor classes. In contrast, the predictions of the proposed method were roughly correct. These results demonstrate that the existing methods tend to bias the major class predictions toward the minor class, while the proposed method does not have such a tendency because it addresses class imbalance.
\par
\section{Limitation}
A limitation of the current trial assumes that the class proportions $(p_1,\ldots,p_C)$ of the unlabeled target samples are roughly close to these of the labeled target samples. Suppose the selected samples have an unexpected bias. In that case, it not only disturbs the inference of class proportions $(p_1,\ldots,p_C)$ but also makes a large difference in the number of labeled samples in each cluster. In future work, we can introduce strategies to automatically select the samples covering the sample variations of the target domain.
\section{Conclusion}
We proposed a semi-supervised domain adaptation method with a novel weakly-supervised clustering pipeline to obtain high-purity clusters even in class-imbalanced cases, common in medical image classification tasks. We formulate cluster-guided domain adaptation for effectively utilizing high-purity clusters in semi-supervised domain adaptation. 
\par
In the digits datasets experiments, we showed that the proposed method effectively improves performance. In addition, we confirmed that the proposed method works well in cervical cancer classification using the real-world WSI dataset. It was also confirmed that the proposed method effectively solved the problem of biased prediction by the classifier trained from a class-imbalanced dataset.
\par
\clearpage
\section{Compliance with Ethical Standards}
This study was performed in line with the principles of the Declaration of Helsinki. Approval was granted by the Ethics Committee of the University of Tokyo and Kyoto University.
\par
\section{Acknowledgments}
This work was supported by JSPS KAKENHI Grant Numbers JP21K19829, JP21K18312, JP21J13083, and JP22H05173.
\par
\bibliographystyle{IEEEbib}
\bibliography{egbib}

\end{document}